\documentclass{article}
\usepackage{ijcai21}

\usepackage{times}

\usepackage{soul}
\usepackage{url}
\usepackage[hidelinks]{hyperref}
\usepackage[utf8]{inputenc}
\usepackage[small]{caption}
\usepackage{graphicx}
\usepackage{amsmath}
\usepackage{booktabs}
\usepackage{subcaption}
\usepackage{color}
\usepackage{multirow}
\usepackage{tabularx}
\usepackage[dvipsnames]{xcolor}
\usepackage[switch]{lineno}

\title{Beyond the Spectrum: Detecting Deepfakes via Re-Synthesis}

\author{
Yang He$^1$,
Ning Yu$^{2,3}$,
Margret Keuper$^{4}$,
Mario Fritz$^1$\\
\affiliations
$^1$CISPA Helmholtz Center for Information Security, Saarbr\"ucken, Germany \\
$^2$Max Planck Institute for Informatics, Saarbr\"ucken, Germany \\
$^3$University of Maryland, College Park, United States \\
$^4$University of Mannheim, Mannheim, Germany \\
\emails
yang.he@cispa.saarland,
ningyu@mpi-inf.mpg.de,
keuper@uni-mannheim.de,
fritz@cispa.saarland
}

\begin{document}

\maketitle

\begin{abstract}
The rapid advances in deep generative models over the past years have led to highly {realistic media, known as deepfakes,} that are commonly indistinguishable from real to human eyes. These advances make assessing the authenticity of visual data increasingly difficult and pose a misinformation threat to the trustworthiness of visual content in general. Although recent work has shown strong detection accuracy of such deepfakes, the success largely relies on identifying frequency artifacts in the generated images, which will not yield a sustainable detection approach as generative models continue evolving and closing the gap to real images. In order to overcome this issue, we propose a novel fake detection that is designed to re-synthesize testing images and extract visual cues for detection. The re-synthesis procedure is flexible, allowing us to incorporate a series of visual tasks - we adopt super-resolution, denoising and colorization as the re-synthesis. We demonstrate the improved effectiveness, cross-GAN generalization, and robustness against perturbations of our approach in a variety of detection scenarios involving multiple generators over CelebA-HQ, FFHQ, and LSUN datasets. Source code is available at \href{https://github.com/SSAW14/BeyondtheSpectrum}{https://github.com/SSAW14/BeyondtheSpectrum}.

\end{abstract}

\section{Introduction}
\label{sec:intro}
\noindent In the past years, image generation and tampering techniques have been evolving quickly, benefiting from the continuous breakthroughs in generative adversarial networks (GANs)~\cite{gan14nips} and its variations~\cite{radford2015dcgan,gulrajani2017wgan,karras2017ProGAN,karras2019StyleGAN,karras2019StyleGAN2,yu2021dual}. The fidelity and diversity of generated images have improved to a level that is arguably already photorealistic. Although fostering the techniques for numerous novel applications~\cite{reed2016generative,thies2016face2face,zhu2017cyclegan,choi2018stargan,yu2019texture,yu2020inclusive,wang2020hijack}, this development, on the other hand, poses new risks as recent results are challenging to be distinguished from real images by human eyes.

Malicious individuals may rely on the above techniques to alter or create the media, and spread misleading information, which will cause unpredictable results. Therefore, several precautions of misuse of the techniques are developed, including targeting the source of forgery~\cite{zhang2020not_my_deepfakes,ning2019iccv_gan_detection,yu2020artificial,yu2020responsible} and detection of forgery, which provides people warning messages to trust the media suitably.

\begin{figure}[!t]
\centering
\includegraphics[trim=0cm 10cm 0cm 0cm, clip=true,width=0.9\linewidth]{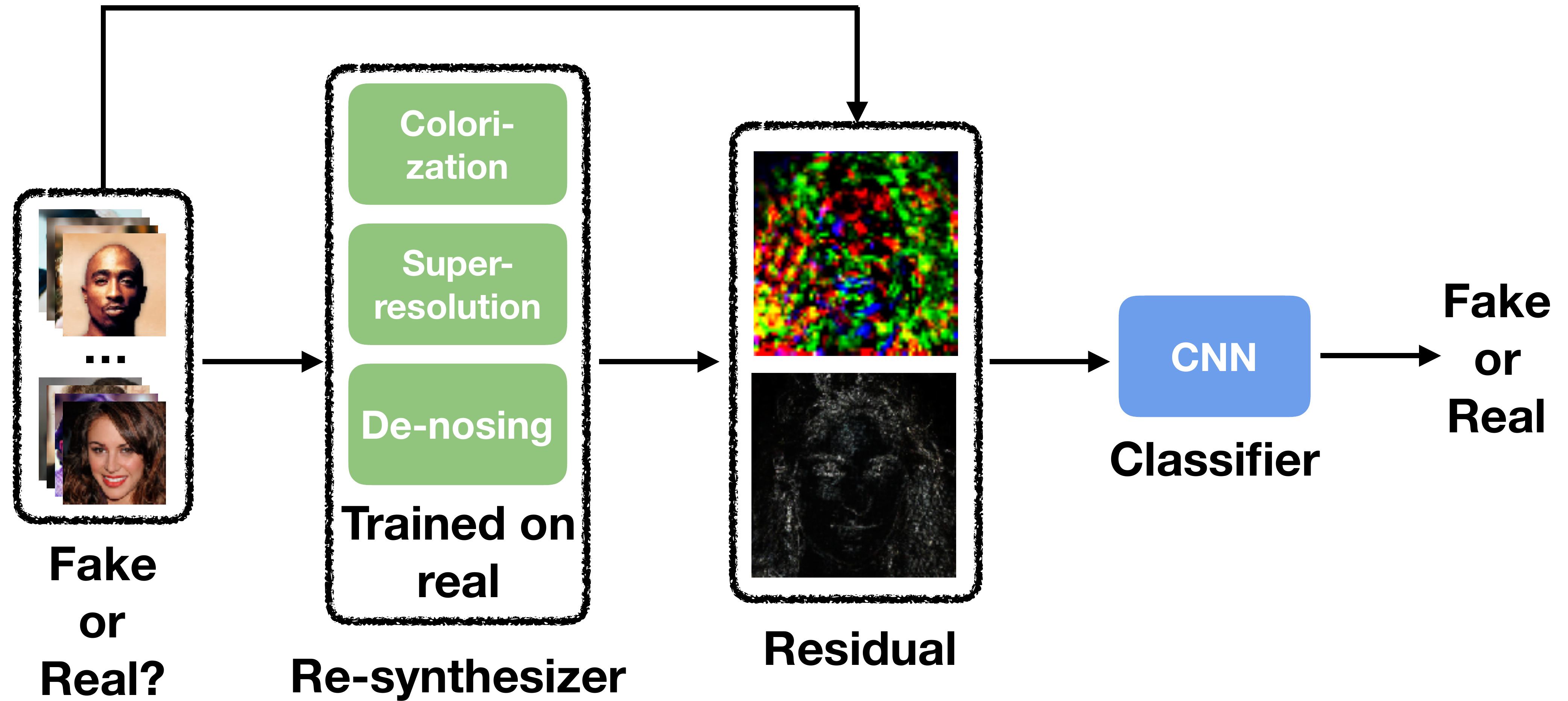}
   \caption{We train a classifier for robust deepfakes detection with an auxiliary re-synthesizer trained on real images which are comprised of several tasks for modeling real image distributions, isolating fake images and extracting robust features in unknown scenarios.  }
\label{fig:teaser_figure_intro}
\end{figure}

In particular, detecting fake images has received tremendous attention, because images are the ubiquitous  media and appear widely in various platforms, such as social networks, advertisement, etc. Besides, detecting fake images is the backbone of many alarm systems working in more sophisticated cases, such as videos. Recent fake detection techniques mainly rely on local regions artifacts~\cite{ning2019iccv_gan_detection,masi2020two_branch,chai2020makes_fake}, global textures~\cite{liu2020texture_fake} or rely on a mismatch in the generated frequency distribution~\cite{durall2019unmasking,zhang2019detecting,frank2020dct2d_detect}. A CNN classifier is typically trained on the extracted features to perform binary classification for fake detection.
However, recent work has shown that such low-level features are rather easy to recognize~\cite{wang2020cvpr_detect} and can be effectively concealed~\cite{margret2020upconvolution,steffen_aaai21}. Due to the steady improvement of generative models and the constantly narrowing gap between real and fake images, this appears not to yield in a reliable and sustainable approach to distinguish real and fake images. Therefore, it motivates us to seek  different, diverse, or at least complementary approaches for robust generated images detection targeting potential unknown configurations.

In order to overcome the issue of excessive reliance on simplistic frequency and low-level artifacts, we propose a novel feature representation for generated images detection. We achieve this by processing \textit{both} real \textit{and} fake images with a generator that in turn induces similar {frequency} artifacts to \textit{both} images while distinct \textit{ residuals}, as sketched in Figure~\ref{fig:teaser_figure_intro}. The generator is trained with real images to perform several synthesis tasks. We aim to complete information for real images from some sketched information, such as colorization, denoising and super-resolution. The frequency artifacts become non-discriminative features and will - as we show - not be used for the detection. Instead, the proposed detection mechanism leverages the features of multi-stage reconstruction errors w.r.t. the re-synthesis model. It turns out to be remarkably effective to distinguish real and fake images - which in addition {is more generalized across different GAN techniques} and more robust against a variety of perturbations that try to conceal fakes.

We highlight the contributions and novelty of our work as follows:(1) We present a novel feature representation  for fake image detection based on re-synthesis, which is based on a super-resolution pipeline that learns a detector agnostic to the previously-used simplistic frequency features. (2) We validate the improvements of our method in terms of detection accuracy, {generalization}, and robustness, compared to prior work in a diverse range of settings.

\section{Related Work}
\label{sec:related}
\subsection{Generative Adversarial Networks (GANs)}
GANs~\cite{gan14nips} have achieved tremendous success in modeling data distributions. The breakthroughs mainly come from the improvements of training strategies~\cite{karras2017ProGAN} or model architectures~\cite{karras2019StyleGAN,karras2019StyleGAN2,yu2021dual}. The current state-of-the-art GANs are capable of producing high-resolution images with realistic details, which make it rather challenging for human eyes to distinguish generated and real images apart. 
Specifically, we study the problem of detecting deepfakes with the state of the art GANs: ProGAN~\cite{karras2017ProGAN}, StarGAN2~\cite{choi2020starganv2}, StyleGAN~\cite{karras2019StyleGAN}, StyleGAN2~\cite{karras2019StyleGAN2}.

\subsection{Low-Level Artifacts produced by GANs}
GANs have been significantly improved in recent years and are able to synthesize high fidelity images fooling human eyes. However, there persist some problems in GANs revealing the differences between generated distributions and real ones because of commonly-used up-convolution (or deconvolution) operation~\cite{margret2020upconvolution}, which maps low-resolution tensors to high-resolution ones. 
Yet these problems are never long-lasting compared to the steady improvement of GANs. For example, spectral regularization is proposed~\cite{margret2020upconvolution} to close the gap in the spectral domain.
Recently, \cite{steffen_aaai21} learn an additional discriminator with spectrum inputs with adversarial training, and the frequency gap of fake images is reduced further.
Hence, it is not sustainable to establish fake detection mechanisms based on the known problems of GANs - these problems are also known to malicious individuals and can be sidestepped along with the steady development of improved GANs. That motivates us to propose a novel mechanism for detecting fake images, which is not reliant on such low-level artifacts.

\subsection{Fake Detection with Spatial Analysis}
Because of the emerging risks of fake information explosion, 
fake image detection has become an increasingly crucial and prevalent topic~\cite{marra2018gan_detect,nataraj2019co_occurrence,roessler2019faceforensics++,wang2020cvpr_detect,marra2019full}. Recent work analyzes different low-level visual pattern representations so as to attribute images into real or fake~\cite{marra2019fingerprints,ning2019iccv_gan_detection,liu2020texture_fake,chai2020makes_fake}. First,~\cite{ning2019iccv_gan_detection,marra2019fingerprints} validate that GAN training naturally leaves a unique fingerprint for each model, which serves as a visual cue for fake detection. Furthermore,~\cite{liu2020texture_fake} design a network to induce texture representations using a Gram matrix, and validate that global textures at different levels of a CNN are effective cues for fake detection. Also, Laplacian of Gaussian (LoG) is augmented along with images to foster fake image and video detection~\cite{masi2020two_branch}. Last, according to a patch-level prediction from different stages of a CNN, it has been shown that hair and background are the most informative areas for detecting fake facial images~\cite{chai2020makes_fake}, which may help detection across various data distribution. In this paper, we compare to GAN fingerprint techniques~\cite{marra2019fingerprints,ning2019iccv_gan_detection} as representatives of this approach, and show improved performance.

\subsection{Fake Detection with Frequency Analysis}
Frequency analysis has a long history and broad applications in image processing. Several recent methods based on analyzing frequency patterns of images are adapted to fake detection. A simple yet effective method based on azimuthally-averaged spectrum magnitude and SVM is proposed~\cite{durall2019unmasking}. The 2d-FFT magnitudes serve as input features for CNN binary classification~\cite{zhang2019detecting,wang2020cvpr_detect}. In a similar spirit, 2d-DCT is also studied as CNN input features~\cite{frank2020dct2d_detect} and demonstrates improved detection results compared to image-based method~\cite{ning2019iccv_gan_detection}. To the best of our knowledge, the most recent state-of-the-art detector leverages global and local 2D DCT features~\cite{qian2020thinking_frequency_fake} and further validates the effectiveness of frequency analysis in terms of detecting fake images. In this paper, we compare to azimuthally-averaged spectrum, 2d-FFT, and 2d-DCT as representatives of this approach, and show improved performance.

\section{Detection by Re-Synthesis}
\label{sec:method}

\begin{figure*}[!t]
\centering
\begin{minipage}{0.68\textwidth}
\includegraphics[width=\linewidth]{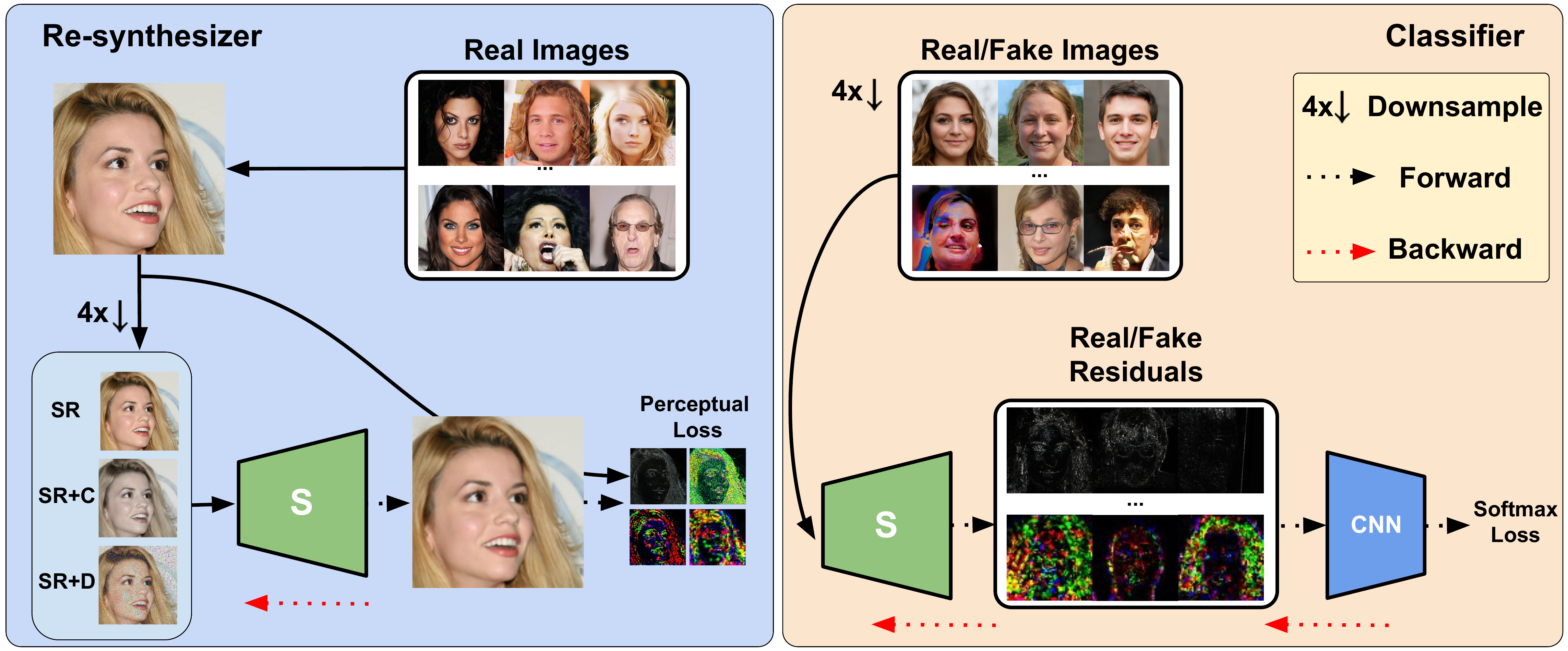}
   \caption{The diagram of our detection pipeline. 
   {Our end-to-end model has two components.  A classifier is trained with real/fake images. We learn a re-synthesizer with real images only to help extracting robust features and isolating fake images. The synthesizer takes different forms of inputs to capture various visual patterns from those tasks for robust representations, including super-resolution (SR), colorization (C) and denoising (D). }}
\label{fig:method_pipeline}
\end{minipage}
\begin{minipage}{0.01\textwidth}
\quad
\end{minipage}
\begin{minipage}{0.3\textwidth}
\fontsize{9pt}{11pt}\selectfont
\begin{tabular}{@{}c@{}c@{}c@{}c@{}c@{}c}
& \multirow{2}{*}{Real} & Pro-  & Style- & Style- \\
&  & GAN  & GAN & GAN2 \\
\rotatebox{90}{ Image} &
   \includegraphics[width=0.22\linewidth]{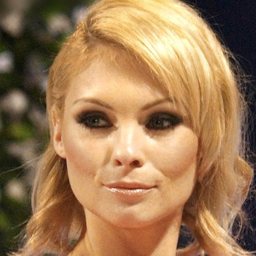} &
  \includegraphics[width=0.22\linewidth]{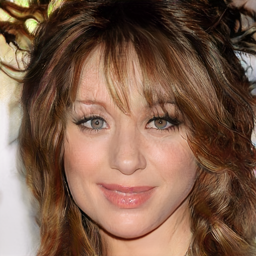} &
  \includegraphics[width=0.22\linewidth]{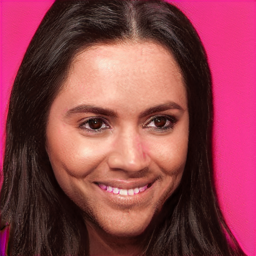} &
  \includegraphics[width=0.22\linewidth]{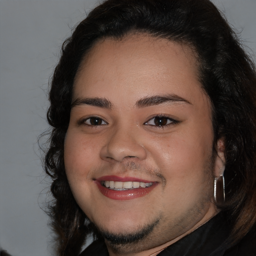}  \\
\rotatebox{90}{ Pixel} &
   \includegraphics[width=0.22\linewidth]{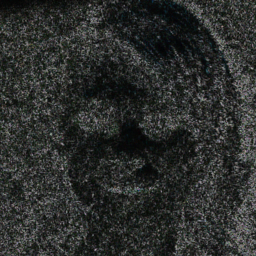} &
   \includegraphics[width=0.22\linewidth]{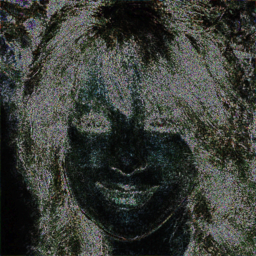} &
   \includegraphics[width=0.22\linewidth]{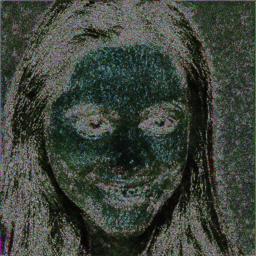} &
   \includegraphics[width=0.22\linewidth]{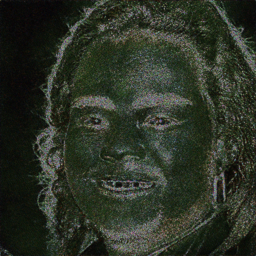} \\
    \rotatebox{90}{ Stage5} &
   \includegraphics[width=0.22\linewidth]{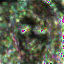} &
   \includegraphics[width=0.22\linewidth]{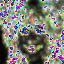} &
   \includegraphics[width=0.22\linewidth]{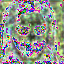} &
   \includegraphics[width=0.22\linewidth]{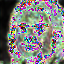}  \\    
   \end{tabular}
   \caption{Visualization of hierarchical structural artifacts for different image sources. The first row indicates query images, the last two rows indicate pixel- and stage5-level artifacts of the perceptual network.}
\label{fig:vis_artifacts}

\end{minipage}
\end{figure*}

The key goal of our work is to propose a more robust feature presentation for fake image detection, which should be generalized enough across different fake sources, robust enough against image perturbations, and more importantly, not reliant on low-level artifacts naturally induced by generative models.

To achieve this, we propose a novel method that is explicitly designed to re-synthesize testing images according to several auxiliary tasks to model the distribution for real images,
for the purpose of robust fake image detection.
Our model has two components, a \textit{re-synthesizer} $\mathbf{S}$ and a \textit{classifier} $\mathbf{C}$, and they are jointly trained end to end.
The re-synthesizer aims to distinct real/fake images that it is only learned with real images to provide different residuals from fake ones. Besides, the re-synthesizer allows us to incorporate a variety of noisy patterns to avoid overfitting on the representations from it for the robust detection in unknown scenarios.
As an instance, our method incorporates a super-resolution model, which aims to predict the high-frequency information from low-resolution inputs. 
Owing to the differences w.r.t high-frequency information between real and fake images,
this allows us to build a fake image detector based on the residual errors (i.e., reconstruction errors) formalized by this super-resolution model. Further, we take different forms of inputs for the super-resolution model to capture richer and more robust visual representations from real images for improving the robustness of detection.

\subsection{{Re-Synthesis Residuals} as Structural Artifacts}
We train a re-synthesizer $\mathbf{S}$ on real images, which takes a downsampled version to reconstruct the original image,
and regard the structural reconstruction error maps as features for classification. Particularly, $\mathbf{S}$ is trained on real images only, and then we are able to show different residual distributions for real and fake images.

Mathematically, given a dataset $\mathcal{D}^{+}$ representing real images, we first train a super-resolution model $\mathbf{\Phi}$ on $\mathcal{D}^{+}$, which is formulated as a regression task with the loss function
\begin{equation}
    L = \|X-\mathbf{\Phi}(\mathbf{\Omega}(X_{\downarrow}))\|_{1}, \\
    \label{eq:l1}
\end{equation}
where $\mathbf{\Omega}$ could be an image degeneration operation, $X\in\mathcal{D}^{+}$ and $X_\downarrow$ is a downsampled version of $X$. As a result, our re-synthesizer $\mathbf{S}(\cdot)=\mathbf{\Phi}(\mathbf{\Omega}(\cdot))$. In this work, $\mathbf{\Omega}=\{I, G, N\}$, referring to the indentity, grayscaling and noising operations respectively.

After training the super-resolution model, we apply the structural artifact with downsampled images $|X-\mathbf{\Phi}(X_{\downarrow})|$ as the feature for fake detection. We collect another dataset of fake images from a generator, denoted as $\mathcal{D}^{-}$. A fake image detector is then trained on $\mathcal{D}^{+}\cup\mathcal{D}^{-}$. We formulate the detector as a neural network classifier $\mathbf{C}(\cdot)$ trained with softmax loss. In testing, given a query image $X^{*}$, the detection decision is formulated as
\begin{equation}
    \mathbf{C}(|X^{*}-\mathbf{\Phi}(X^{*}_{\downarrow})|). \\
    \label{eq:detect}
\end{equation}

\subsection{Hierarchical Artifacts via Perceptual Loss}
\label{subsec:visual_cues}
Despite the success of low-level artifacts in fake detection~\cite{ning2019iccv_gan_detection,masi2020two_branch}, high-level information remains unexplored and should be equally effective, such as semantic parts, global coherence, etc. Perceptual loss~\cite{johnson2016perceptual}, on the other hand, evaluates the difference between two examples w.r.t CNN activations at various layers, which correspond to hierarchical representations of visual information. 
Consequently, in order to boost our detection pipeline with high-level information, we take advantage of perceptual loss from a pretrained network to train our super-resolution model, instead of only considering the reconstruction errors in pixels.

Let $\mathbf{\Theta}$ be a pretrained network, and $\mathbf{\Theta}_{i}(\cdot)$ be the operation to extract features in the $i$-th stage of a total of $n$ stages. The loss function with perceptions is formulated as

\begin{equation}
    L = \alpha_{0}\|X-\mathbf{\Phi}(X_{\downarrow})\|_{1}+\sum_{i=1}^{n}\alpha_{i}\|\mathbf{\Theta}_{i}(X)-\mathbf{\Theta}_{i}(\mathbf{\Phi}(X_{\downarrow}))\|_{1}, \\
    \label{eq:perceptual_l1}
\end{equation}
where $\alpha_0,\alpha_1,...,\alpha_n$ are loss weights to control the importance of corresponding feature stages during training. After training the super-resolution model, we leverage the $\ell_1$ residual map at each stage, 
as the feature for fake detection. Notably, we can detect fake images by a single classifier or by a combination of classifiers at different stages. Let $\{\mathbf{C}_0(\cdot)$, $\mathbf{C}_1(\cdot)$, ..., $\mathbf{C}_n(\cdot)\}$ be the set of classifiers, where we define $\mathbf{C}_0(\cdot)$ is the pixel-level classifier, and others are the classifiers trained on artifacts at different stages from the perceptual loss. 
The final decision for input image $X^{*}$ is computed with weights $\beta_0,\beta_1,...,\beta_n$ and formulated as
\begin{equation}
    \beta_{0}\mathbf{C}_0(|X^{*}-\mathbf{\Phi}(X^{*}_{\downarrow})|)+\sum_{i=1}^{n}\beta_{i}\mathbf{C}_i(|\mathbf{\Theta}_{i}(X^{*})-\mathbf{\Theta}_{i}(\mathbf{\Phi}(X^{*}_{\downarrow}))|). 
\label{eq:detect_perceptual}
\end{equation}

\begin{figure}[!t]
\begin{center}
\begin{subfigure}[b]{0.22\textwidth}
\includegraphics[height=0.75\linewidth,width=0.95\linewidth]{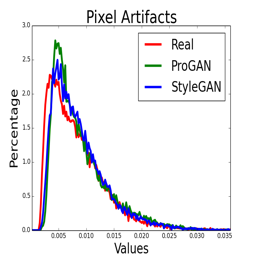}
\end{subfigure}
\begin{subfigure}[b]{0.22\textwidth}
\includegraphics[height=0.75\linewidth,width=0.95\linewidth]{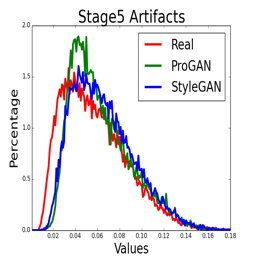}
\end{subfigure}
   \end{center}
   \caption{Histograms of the spatially-average amounts of hierarchical structural artifacts on CelebA-HQ. We observe clear margins between distributions of real and fake (ProGAN or StyleGAN).}
\label{fig:artifacts_distribution}
\end{figure}

In Figure~\ref{fig:vis_artifacts}, we show examples from CelebA-HQ~\cite{karras2017ProGAN} and their structural artifacts in the pixel level and in stage5 of the perceptual network. We observe:
(1) The magnitudes of artifacts of fake images are larger than those of real images, which lay the foundation to distinguish fake from real. In particular, there are more severe artifacts on hairs, eyes, or mouths, which is consistent with the recent study of generalization of fake detection~\cite{chai2020makes_fake}.
(2) The artifact structures are distinct between real and fake, where they look more randomly distributed in real images while with stronger patterns in fake images.
(3) The stage5 perceptual artifacts are more discriminative than pixel artifacts to attribute fake, even between ProGAN and StyleGAN trained on the same dataset. This results in the potential for cross-GAN fake detection.
For quantitative demonstration, in Figure~\ref{fig:artifacts_distribution} we plot the histograms of spatially-averaged amounts of artifacts in the pixel level and in stage5, and show the clear margins between distributions of real and fake.

\section{Experiments}
\label{sec:exp}

\begin{table*}[t]
\fontsize{9pt}{11pt}\selectfont
\resizebox{\linewidth}{!}{
\begin{tabular}{l|c|ccccc|ccccc|ccccc|ccccc|c}
 \toprule
     \multirow{2}{*}{Method} & \multirow{2}{*}{$\mathbf{S}$}& \multicolumn{5}{c}{ProGAN}& \multicolumn{5}{c}{StyleGAN} & \multicolumn{5}{c}{ProGAN$\rightarrow$StyleGAN}& \multicolumn{5}{c}{StyleGAN$\rightarrow$ProGAN} & \multirow{2}{*}{Avg}\\
     &  & Raw & +R & +E & +A & +P & Raw & +R & +E & +A & +P & Raw & +R & +E & +A & +P & Raw & +R & +E & +A & +P\\
    \midrule
      PRNU & \multirow{6}{*}{/}  & 78.3 & 57.1 & 63.5 & 53.2 & 51.3 &  76.5 & 68.8 & 75.2 & 63.0 & 61.9 & 47.4 & 44.8 & 45.3 & 44.2 & 48.9 & 48.0 &  55.1 & 53.6 & 51.1 & 53.6  &  57.3\\
      Image  &  & \underline{99.9} & 58.0 & \underline{99.9} & 56.7 & \textbf{78.8} &  \underline{99.9} & 83.9   & \underline{99.9} & 72.3 & \textbf{81.3} & 51.6 & 50.9 & 51.7. & 51.6 & 51.1 &  52.8 &  50.5  & 54.4 &  51.5 & 54.2 & 67.6\\
      1D Spectrum  &  & 97.5 & 69.6  & 75.7 & 70.0 & 54.9 &  93.0 & 49.1 & 68.2 & 52.2  & 54.4 & 93.0 & 48.6 & 71.6 & 49.6 & 53.5 & 97.5 & 65.0  & 56.9 & 66.1 & 52.3 & 67.9\\
     FFT-2d  & & \underline{99.9}  & 95.9   & 81.8 & \textbf{99.9} & 59.8 & \textbf{100}  & 90.8  & 72.0 & 99.4 & 57.7  & 98.9  & \underline{99.8}   & 63.2 & 61.1 & 56.8 &  77.5  & 54.6  & 56.5 & 76.5 & 55.5 & 78.1\\
     DCT-2d  &  & \underline{99.9} & \underline{99.9} & \underline{99.9} & \textbf{99.9} & 54.4 &   \underline{99.9} & \underline{99.8} & \underline{99.9} & \underline{99.8} & 56.0 & 98.6 & \textbf{99.9} & 98.4 & 81.8  & 54.2 & \underline{99.0} & 95.6 & 98.6 & 97.1 & 55.0 &  89.2\\
     GramNet  & & 100 & 77.1  & 100 & 77.7  & 69.0 & 100 & 96.3 & 100 & 96.3  & 73.3 & 64.0 & 57.3 & 63.7 & 50.9  & 57.1 & 63.1 & 56.4 & 63.8 & 66.8 & 56.2 & 74.6\\
     \hline
     Ours (Pix) & & \textbf{100}   & 99.7   & \underline{99.9} & 99.3. & 57.5 &   \textbf{100}  &  70.0   & \textbf{100} & 97.9 & 57.4 & \underline{99.8}   & 78.6   & \textbf{99.9} & 98.4  & 55.8 & {98.2}  & {99.6}  & {99.2} & 98.1 & 55.0 & 87.0\\
     Ours (Stage5)  & SR & \underline{99.9} &  \underline{99.9}  & \underline{99.9} & \textbf{99.9}  & {71.7} & \textbf{100}  & \underline{99.8}  & \textbf{100} & \underline{99.8} & {68.4} & {99.4}  & 99.5 & {99.7} & 99.4 & \underline{70.7}  &  96.0 &  97.9  & 97.1 & \underline{98.7} & \underline{67.4} & 92.5\\
     Ours (Avg) & & \textbf{100} & \textbf{100} & \textbf{100} &  99.7 & 64.5 & \textbf{100} & {98.7} & \textbf{100} & \textbf{99.9} & 66.7 & \textbf{100} & 97.8 & \textbf{99.9} & \underline{99.8} & {67.0} &  \textbf{99.5} & \textbf{99.9} & \underline{99.8} & \textbf{100} & {66.1} & 92.2\\
     
     \hline
     Ours (Pix) & & \underline{99.9}   & 99.8   & \underline{99.9} & 99.4 & 55.3 &   \textbf{100}  &  75.1   & \textbf{100} & 98.5 & 59.5 & \underline{99.8}   & 80.8   & \textbf{99.9} & 98.4  & 55.9 & 99.0  & 99.5  & {99.2} & 98.3 & 55.1 & 88.7\\
     Ours (Stage5)  & SR+C & \underline{99.9} &  \underline{99.9}  & \underline{99.9} & \textbf{99.9}  & {72.5} & \textbf{100}  & \underline{99.8}  & \textbf{100} & 99.7 & {69.2} & {99.4}  & \textbf{99.9} & {99.7} & \textbf{99.9} & \textbf{72.5}  &  97.2 &  98.4  & 98.6 & 98.6 &  \textbf{68.0}   & \textbf{93.7} \\
     Ours (Avg) & & \textbf{100} & \textbf{100} & \textbf{100} &  99.7 & 64.5 & \textbf{100} & {99.5} & \textbf{100} & \textbf{99.9} & 68.5 & \textbf{100} & 97.8 & \textbf{99.9} & \underline{99.8} & \underline{68.5} &  {99.3} & \textbf{99.9} & \textbf{100} & \textbf{100} & {66.1} & 93.2\\
     \hline
     Ours (Pix) & & \textbf{100}   & 99.8   & \underline{99.9} & 99.3. & 58.2 &   \textbf{100}  &  77.3   & \textbf{100} & 99.2 & 59.9 & {99.7}   & 79.0   & \textbf{99.9} & 98.2  & 57.2 & {99.2}  & \underline{99.8}  & {99.2} & 98.1 & 55.0 & 88.9\\
     Ours (Stage5)  & SR+D & \underline{99.9} &  \underline{99.9}  & \underline{99.9} & \textbf{99.9}  & \underline{73.0} & \textbf{100}  & \textbf{99.9}  & \textbf{100} & 99.7 & \underline{73.4} & {99.2}  & 99.4 & \underline{99.8} & 99.4 & 69.5  &  95.2 &  97.6  & 97.1 & 98.2 & {66.8} & \underline{93.4}\\
     Ours (Avg) & & \textbf{100} & \textbf{100} & \textbf{100} &  99.7 & 66.5 & \textbf{100} & {98.7} & \textbf{100} & \textbf{99.9} & 68.8 & \textbf{100} & 97.6 & \textbf{99.9} & \underline{99.8} & {67.2} &  {98.0} & \textbf{99.9} & \underline{99.8} & \textbf{100} & {65.2} & 93.1\\
    \bottomrule

\end{tabular}}
\caption{Classification accuracy ($\%$) on CelebA-HQ. The models are trained with fake images from either ProGAN or StyleGAN, and tested on a variety of settings including spectrum regularization (+R), spectrum equalization (+E), spectral-aware adversarial training (+A) and an ordered combination of image perturbations (+P). For our models, we apply 3 re-synthesizers  including super-resolution (SR), super-resolution combined with colorization (SR+C) and super-resolution combined with denoising (SR+D).}
\label{tab:compare_results_celebahq}
\end{table*}

\begin{figure}[!t]
\begin{center}
\begin{tabular}{@{}c@{}c@{}c@{}c@{}c@{}c}
\rotatebox{90}{ Official} &
   \includegraphics[width=0.22\linewidth]{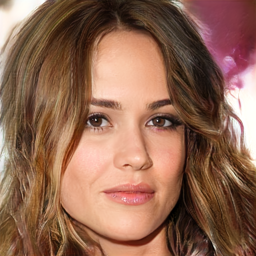} &
   \includegraphics[width=0.22\linewidth]{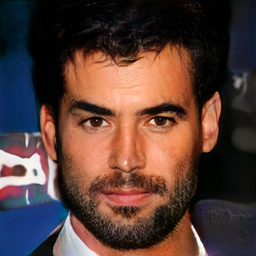} &
   \includegraphics[width=0.22\linewidth]{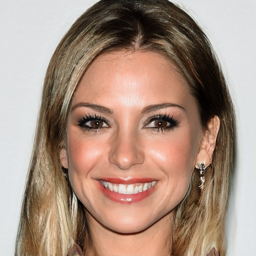} &
   \includegraphics[width=0.22\linewidth]{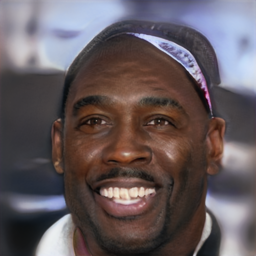} \\
   \rotatebox{90}{ +R } &
   \includegraphics[width=0.22\linewidth]{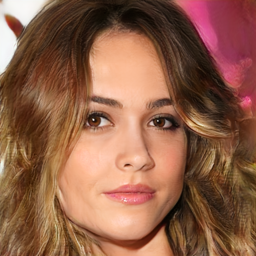} &
   \includegraphics[width=0.22\linewidth]{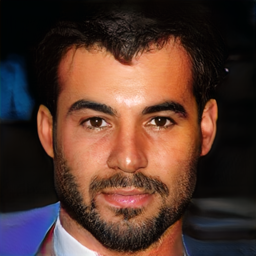} &
   \includegraphics[width=0.22\linewidth]{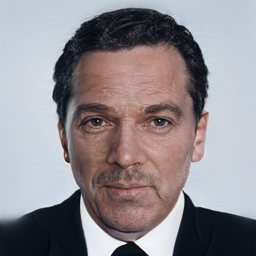} &
   \includegraphics[width=0.22\linewidth]{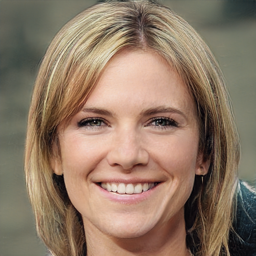}  \\
     \rotatebox{90}{ +A } &
   \includegraphics[width=0.22\linewidth]{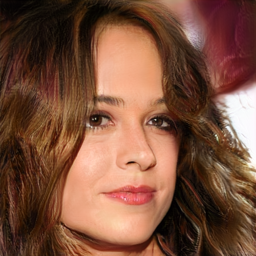} &
   \includegraphics[width=0.22\linewidth]{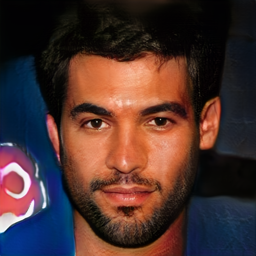} &
   \includegraphics[width=0.22\linewidth]{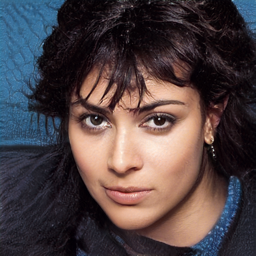} &
   \includegraphics[width=0.22\linewidth]{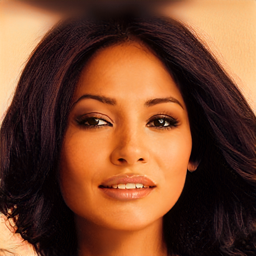}  \\
   & \multicolumn{2}{c}{ ProGAN} & \multicolumn{2}{c}{ StyleGAN} \\
   \end{tabular}
   \end{center}
   \caption{Testing fake image examples from the official ProGAN, StyleGAN, as well as the generators with regularization (+R)~\protect\cite{margret2020upconvolution} and spectral-based adversarial training (+A)~\protect\cite{steffen_aaai21}. We test the detectors on different GAN models for the comparison of robustness.}
\label{fig:vis_example_spectrum}
\end{figure}

We present the experimental setup and configuration details from section~\ref{subsec:exp_datasets} to~\ref{subsec:implementation}. Accordingly, we provide the results and discussion in the rest of this section.

\subsection{Datasets}
\label{subsec:exp_datasets}
\paragraph{CelebA-HQ} is a 1024$\times$1024 facial image dataset released by~\cite{karras2017ProGAN}. We choose to necessarily conduct experiments on faces because of the prevalent applications of identity recognition and biometric feature recognition. From this dataset, we prepare 25k/25k real and fake images as the training set, and 2.5k/2.5k images as the testing set.

\paragraph{FFHQ}~\cite{karras2019StyleGAN} provides another 70k 1024$\times$1024 facial images. Compared to CelebA-HQ, it covers
vastly more variation on age, ethnicity, eyeglasses, hats, background, etc. To perform cross-domain detection, we test on this dataset all the detectors trained on CelebA-HQ.

\paragraph{LSUN} is large-scale scene dataset~\cite{yu15lsun}. We select bedroom and church\_outdoor classes to perform fake detection beyond human faces. We prepare 50k/50k real and fake images for training, and 10k/10k images for testing.

\subsection{Deep Fake Detection Methods}
We compare our method to the recently-proposed methods, including PRNU~\cite{marra2019fingerprints}, plain image~\cite{ning2019iccv_gan_detection}, 1d spectrum~\cite{durall2019unmasking}, FFT-2d magnitude~\cite{wang2020cvpr_detect,zhang2019detecting}, DCT-2d~\cite{frank2020dct2d_detect} and GramNet~\cite{liu2020texture_fake}.
For our models, we report the performance based on the pixel- and stage5-level artifacts using ResNet-50~\cite{he2016resnet}, which capture low- and high-level spatial errors respectively. In addition, we also report the performance of averaging between the two artifacts. In particular, we compare three re-synthesizers in our study: (1) A super-resolution model is used which predicts high-resolution images from downsampled images, namely \textbf{SR} for short; (2) A super-resolution model which takes partially gray-scaled images to predict high-resolution color images , referred as \textbf{SR+C}. $50\%$ images are performed grayscaling operation. For those images, $10\%\sim25\%$ pixels are randomly set to gray-scale versions. (3) A super-resolution model which takes noisy images to predict high-resolution clean images, referred as \textbf{SR+D}. $50\%$ images are performed noising operation and Gaussian noises with standard deviation 4/255 are applied.

\subsection{Robustness against Image Perturbations}
Image perturbations are able to alter image details or distributions while preserving the contents of them (e.g., denoising, JPEG compression, etc), which may be used to process fake images and challenge the detectors. In order to compare the robustness of different methods, we test on perturbed fake images that are not seen in training phases. First, we calibrate the frequency distributions of generated images using the equalization operation~\cite{margret2020upconvolution}. We regard this operation as post-processing to reduce the frequency distribution gap between real and fake images. Second, we follow the protocol in the previous work~\cite{ning2019iccv_gan_detection} to perturb testing images with an ordered combination of JPEG compression, blurring, cropping, and adding Gaussian noise.

\begin{figure}[!t]
\begin{center}
\begin{tabular}{@{}c@{}c}
\includegraphics[width=0.45\linewidth]{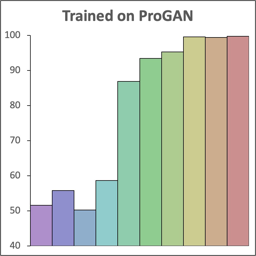} &
\includegraphics[width=0.45\linewidth]{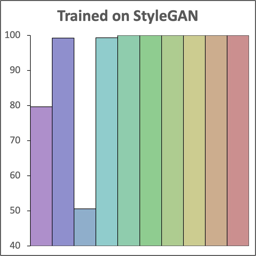} \\
\multicolumn{2}{l}{ \includegraphics[width=\linewidth]{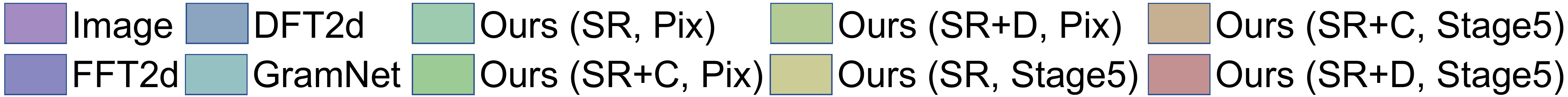} } \\
\end{tabular}
   \end{center}
   \caption{Classification accuracy (\%) on StarGAN2 (CelebA-HQ). We employ detectors trained on ProGAN or StyleGAN. The robustness of our models is clearly observed w.r.t cross-domain detection.}
\label{fig:stargan2}
\end{figure}

\subsection{Generators}
We generate fake images from ProGAN~\cite{karras2017ProGAN}, StarGAN2~\cite{choi2020starganv2}, StyleGAN~\cite{karras2019StyleGAN}, and StyleGAN2~\cite{karras2019StyleGAN2}.
To overcome the known frequency artifacts of up-convolution operations in the generators and provide unknown detection scenarios, we apply the spectral regularization (+R)~\cite{margret2020upconvolution} and spectral-aware adversarial training (+A)~\cite{steffen_aaai21} to finetune released GANs and Figure~\ref{fig:vis_example_spectrum} shows some examples. Besides, we also apply spectrum equalization (+E), which is similar to the regularization, but we process the fake images as a postprocessing.

\subsection{Implementation Details}
\label{subsec:implementation}
We train the super-resolution model on the real images from CelebA-HQ or LSUN. We build a 4$\times$ super-resolution model using~\cite{zhang2018RDN} supervised by the $ell_1$ pixel loss plus VGG-based perceptual loss~\cite{johnson2016perceptual,wang2018hdpix2pix}.  We set the loss weight for each feature from pixel to stage 5 as [1, 1/32, 1/16, 1/8, 1/4, 1].
Second, we train the detectors on pixel artifacts and stage5 artifacts from the VGG network. For each detector, we train a ResNet-50~\protect\cite{he2016resnet} from scratch for 20 epochs using the SGD optimizer with momentum. The initial learning rate is $0.01$ and we reduce it to $0.001$ at the $10$th epoch.

\subsection{Results on CelebA-HQ}
\label{subsec:exp_celeba_hq}
We conduct experiments on CelebA-HQ and compare our method to the baselines, as listed in Table~\ref{tab:compare_results_celebahq}. In specific, our model (Avg) leverages the hierarchical structural artifacts in a combination mode as described in Eq.~(\ref{eq:detect_perceptual}), where we set (1/2 , 1/2) as the weights for combining the final scores of the pixel and stage 5 artifacts. 
In Table~\ref{tab:compare_results_celebahq}, we present the results including the training and testing images are from the same generator and the images are from the different generator. For each part, we not only test on the raw images, but also test on the different challenges of spectrum regularization (+R), equalization (+E), adversarial training (+A) and a combination of perturbations (+P), as discussed before.

\begin{table}[t]
\fontsize{9pt}{11pt}\selectfont
\centering
\resizebox{0.95\linewidth}{!}{
\begin{tabular}{l|c|ccc|ccc|c}
 \toprule
     \multirow{2}{*}{Method} & \multirow{2}{*}{$\mathbf{S}$}  & \multicolumn{3}{c}{ProGAN}& \multicolumn{3}{c}{StyleGAN} & \multirow{2}{*}{Avg}  \\
     &  & SG & SG2$^{1}$ & SG2$^{2}$ & SG & SG2$^{1}$ & SG2$^{2}$  \\
    \cmidrule(lr){1-9}
       PRNU  &  \multirow{6}{*}{/}  & 46.3  & 44.5 &  46.1 & 63.2 & 53.7 & 43.5 & 45.4 \\
       Image   & & 49.8 &   45.2   & 44.9 & 50.0 & 48.0 & 49.2 & 47.9  \\
       1D Spectrum  & &  51.7& 54.5&  51.4 & 50.9 & 61.3 & 53.7 & 53.9  \\
       FFT-2d  &  &  50.7  & 58.4 &  {72.0} & 56.1  & {94.5} & \underline{95.2} & 71.1 \\
       DCT-2d & &  \textbf{87.3}  &   62.4   & 67.3 & \textbf{88.8} &  93.8  & 93.6  & 82.2      \\
       GramNet & &  50.1   & 45.6 & 45.7 &  66.2  & 46.8 & 48.5 & 50.5  \\
      \hline
       Ours (Pix) &  &  70.6 &   {81.2} &     {91.3} & 60.2 & {95.6} & {96.2}& 82.5  \\
       Ours (Stage5)   & SR & {83.0} &  54.0 &  55.1 & \underline{86.2}  & 71.3 & 68.3 & 69.7  \\
       Ours (Avg) &  & {79.1} &  {62.1} &  70.9 & {77.8}  & 89.7 & 90.0 & 78.3 \\
      \hline
       Ours (Pix) &  &  71.6 &   \textbf{93.4} &  \underline{93.8} & 64.6 & \underline{97.2} & \underline{96.9}  & 86.3 \\
       Ours (Stage5)   & SR+C &  \underline{83.1} &  54.1  &     55.8 &  85.5 & 71.5  &  68.4  & 69.7  \\
       Ours (Avg) &  & 81.2  & 88.2   &  88.5   & 80.3  & 91.5  & 92.0 & \underline{87.0}   \\
      \hline
       Ours (Pix) & \multirow{3}{*}{SR+D} & 71.0 & \textbf{93.4} & \textbf{94.1} & 65.4  & \textbf{97.4} & \textbf{97.0} & 86.4 \\
       Ours (Stage5)   & & 81.8  & 53.5 & 54.6 &    83.7 & 66.8 &  65.7  & 67.7  \\
       Ours (Avg) &  & 78.8  & \underline{88.8}   & 89.6    &  79.7 & 93.9   & 93.4  & \textbf{87.4} \\
    \bottomrule

\end{tabular}}
\caption{Classification accuracy ($\%$) w.r.t cross-domain detection on FFHQ. Classification models are trained on CelebA-HQ with ProGAN or StyleGAN. Fake images from StyleGAN and StyleGAN2 trained on FFHQ are used for testing, abbreviated as SG and SG2. For StyleGAN2, we generate images with psi value of 0.5 or 1.0, referred as SG2$^{1}$ and SG2$^{2}$. }
\label{tab:compare_results_acc_cross_detect}
\end{table}

First,  we observe our detectors achieve remarkable results in average. We emphasize:
(1) Spectrum-based methods~\cite{durall2019unmasking,zhang2019detecting} achieves decent performance and robustness in the cross-GAN settings. For example, the performance of~\cite{durall2019unmasking} and~\cite{zhang2019detecting} deteriorates only by ~10\% on average in the cross-GAN settings. 
(2) DCT-2d-based method is robust across domains and robust against processing on frequency. However, it is sensitive to the image perturbations, deteriorating the performance severely.
(3) In contrast, our model with any of the features achieves better results compared to the baselines. Especially the deterioration is imperceptible when applying spectral processing or testing cross domains.
(4) Introducing colorization or denoising into the re-synthesis helps learning more robust features and achieving more favorable results.

In addition to our advantageous performance, we also point out a few insightful discoveries as follows:
(1) Our stage5-based detector obtains the best accuracy in the cross-domain settings with image perturbations, which validates the robustness of leveraging high-level information.
(2) Our average detector achieves better performance than our other two detectors in many tests, indicating the beneficial synergy between our pixel artifacts and stage 5 artifacts.  We reason this as the synergy effect between pixel and stage artifacts. Therefore, we suggest considering more about high-level cues for fake detection, instead of looking at the local only.

\begin{figure}[t]
\large
\begin{center}
\begin{tabular}{@{}c@{}c@{}c@{}c@{}c@{}c@{}c@{}c@{}c}
&  & {\footnotesize Fake Image} & & {\footnotesize Image-based} & &  {\footnotesize Pix} & &  {\footnotesize Stage5} \\
   \rotatebox{90}{ \footnotesize  ProGAN} & &
   \includegraphics[width=0.24\linewidth]{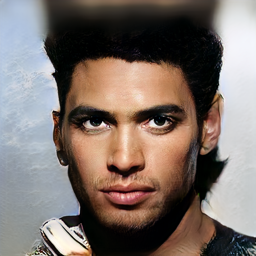} & \text{ } &
   \includegraphics[width=0.24\linewidth]{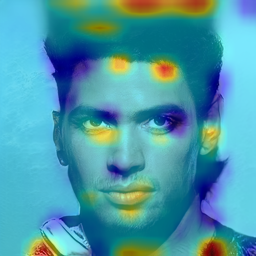} & \text{ } &
   \includegraphics[width=0.24\linewidth]{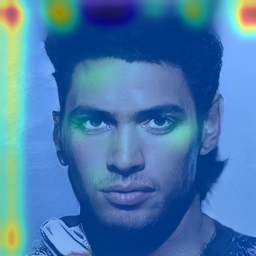} & \text{ } &
   \includegraphics[width=0.24\linewidth]{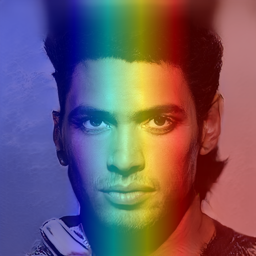} \\   
    \rotatebox{90}{ \footnotesize  ProGAN+R} & &
   \includegraphics[width=0.24\linewidth]{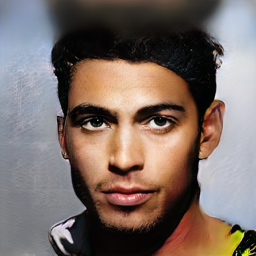} & \text{ } &
   \includegraphics[width=0.24\linewidth]{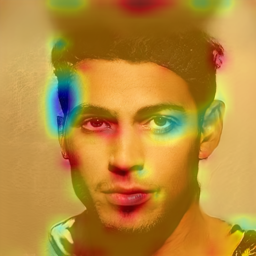} & \text{ } &
   \includegraphics[width=0.24\linewidth]{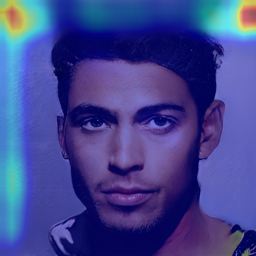} & \text{ } &
   \includegraphics[width=0.24\linewidth]{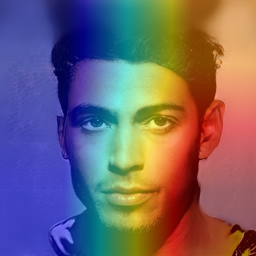} \\       
   \end{tabular}
   \end{center}
   \caption{Visualization of class activation maps for recognizing as fake images in CelebA-HQ. We compare our methods (SR) based on  pixel and stage5 artifacts with image-based approach~\protect\cite{ning2019iccv_gan_detection}.
   }
\label{fig:cam_artifacts_sr}
\end{figure}

Additionally, we also test the above detectors on recent proposed StarGAN2, where the results are visualized in Figure~\ref{fig:stargan2}. We highlight several points:
(1) Most competing detectors fail to recognize StarGAN2 generated images owing to the domain shift. Even though DCT2d detector perform well in Table~\ref{tab:compare_results_celebahq}, it hardly recognizes the fakeness from StarGAN2 images.
(2) All of our detectors with different synthesis modules and visual features achieve accuracy higher than 85\% and most of them are close to 100\%. Besides, the improved results are achieved for ProGAN after employing colorization or denoising in the re-synthesis, which shows the effectiveness of introducing noises into the re-synthesizer.

{To further demonstrate the robustness of our detectors, we visualize the class activation maps (CAMs) for our methods (SR) and previous detectors~\cite{ning2019iccv_gan_detection} in Figure~\ref{fig:cam_artifacts_sr}. From this figure, we are able to observe the CAMs of our approach are quite stable for similar input fake images, even though spectral regularization~\cite{margret2020upconvolution} is applied to reduce the spectrum distrbution between real and fake images. In contrast, the image based detector~\cite{ning2019iccv_gan_detection} is sensitive to the changes and outputs an opposite highlighted region. Therefore, we reason performance reduction for previous approaches to their sensitivities.  }

\subsection{Results on FFHQ}
We conduct experiments for cross-domain detection with FFHQ, where the results are listed in Table~\ref{tab:compare_results_acc_cross_detect}. In the experiments, we do not train additional detectors; instead directly test the models on novel data, which are trained on CelebA-HQ. We apply the real images from FFHQ and fake images from  StyleGAN and StyleGAN2 trained on FFHQ. Because this setting already challenges most detectors, we do not employ additional perturbations as CelebA-HQ. 
From Table~\ref{tab:compare_results_acc_cross_detect} we observe:
(1) Our model using pixel inputs achieves the best performance, and outperforms the DCT-2d based approach when StyleGAN2 is tested. 
(2) Our stage5-based models achieve comparable performance to DCT-2d and are better than our pixel-based when StyleGAN is tested, which further shows the complementary capability of low- and high-level artifacts to deal with wider detection scenarios.
(3) We can observe clear improvements when colorization or denoising is applied, indicating more robust features are obtained and thus achieving better results in cross-datasets detection.  We owe this observation to the avoid of overfitting on the training data by the noisy inputs of our synthesizers.

\begin{table}[t]
\fontsize{9pt}{11pt}\selectfont
\centering
\resizebox{0.9\linewidth}{!}{
\begin{tabular}{l|c|p{1.1cm}<{\centering}p{1.1cm}<{\centering}|p{1.1cm}<{\centering}p{1.1cm}<{\centering}|c}
 \toprule
     \multirow{2}{*}{Method} & \multirow{2}{*}{$\mathbf{S}$}  &\multicolumn{2}{c}{Bedroom}& \multicolumn{2}{c}{Church\_Outdoor} & \multirow{2}{*}{Avg} \\
     & & Raw & +P & Raw & +P  &  \\
    \cmidrule(lr){1-7}
      Image     &  \multirow{4}{*}{/}   & \textbf{100} & 54.2 & \textbf{99.9} & 57.7 & 78.0   \\
      FFT-2d   &  & 99.6 & 61.5 & \underline{99.8} & 57.5 & 79.6   \\
      DCT-2d  &  & \underline{99.9}  & 59.5 &  \textbf{99.9} & 60.5 & {80.0}  \\
      GramNet & & \underline{99.9}  & {73.7} & 99.9 & 68.7 & {85.5}      \\
     \hline
      Ours (Pix) & &  \underline{99.9} & 56.6 &  \underline{99.8}  & 58.4 & 78.7    \\
       Ours (Stage5) & SR & 99.6  & \textbf{76.5} &  99.5  & {75.2}  & {87.7}     \\
      Ours (Avg) & & \underline{99.9}  & {65.1}  &  \textbf{99.9} & {69.9} &  83.7    \\      
     \hline
      Ours (Pix) & \multirow{3}{*}{SR+C} &  \underline{99.9} & 57.5 &  \underline{99.8}  & 56.4 & 78.4    \\
       Ours (Stage5) & & 99.6  & \underline{76.1} &  99.0  & \textbf{96.0}  & \textbf{92.7}     \\
      Ours (Avg) & & \underline{99.9} & 71.0 & \underline{99.8} & 73.0 & 86.0    \\     
     \hline
      Ours (Pix) & \multirow{3}{*}{SR+D} &  \underline{99.9} &  56.9  & 99.7  & 57.0 &  78.4   \\
       Ours (Stage5) &  & 99.2  & 75.9 & 98.8  &  \underline{95.9} &  \underline{92.2}    \\
      Ours (Avg) & & \underline{99.9} & 71.4 & \underline{99.8}  & 72.9  &  86.0   \\ 
    \bottomrule
\end{tabular}}
\caption{Classification accuracy ($\%$) on LSUN datasets. We test on the fake images from official released models (Raw) and employ a combination of perturbations on them (+P). }
\label{tab:compare_results_lsun}
\end{table}

\subsection{Results on LSUN}
\label{subsec:exp_lsun}
\begin{figure}[t]
\large
\begin{center}
\begin{tabular}{@{}c@{}c@{}c@{}c@{}c@{}c@{}c@{}c}
 & {\footnotesize Fake Image} & &  {\footnotesize Image-based} & &  {\footnotesize  Pix} & &   {\footnotesize  Stage5} \\
    &
   \includegraphics[width=0.24\linewidth]{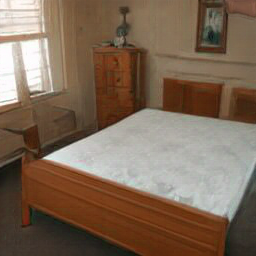} & \text{ } &
   \includegraphics[width=0.24\linewidth]{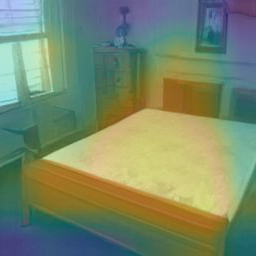} & \text{ } &
   \includegraphics[width=0.24\linewidth]{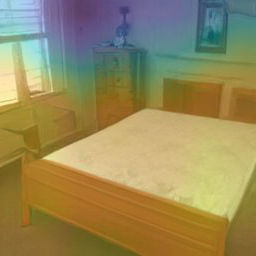} & \text{ } &
   \includegraphics[width=0.24\linewidth]{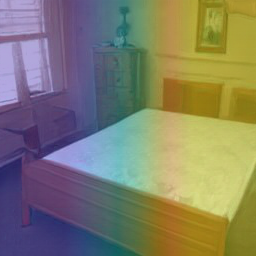} \\   
    &
   \includegraphics[width=0.24\linewidth]{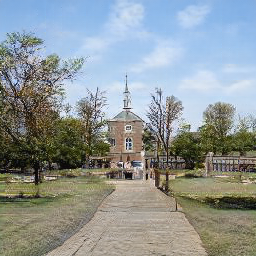} & \text{ } &
   \includegraphics[width=0.24\linewidth]{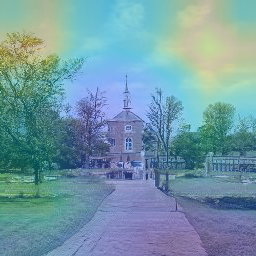} & \text{ } &
   \includegraphics[width=0.24\linewidth]{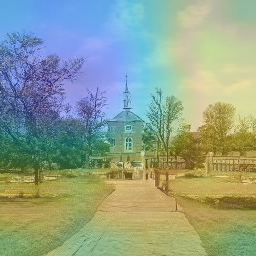} & \text{ } &
   \includegraphics[width=0.24\linewidth]{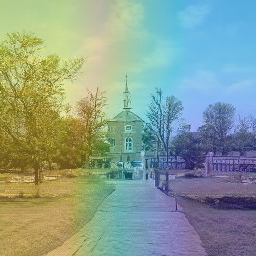} \\   
   
   \end{tabular}
   \end{center}
   \caption{Visualization of class activation maps for recognizing as fake images in LSUN. We compare our methods (SR) based on the artifacts of pixel and stage5 with image-based approach~\protect\cite{ning2019iccv_gan_detection}.
   }
\label{fig:cam_artifacts_bedroom}
\end{figure}

We use official released StyleGAN for generating images of bedroom and StyleGAN2 for generating images of church\_outdoor. In Table~\ref{tab:compare_results_lsun}, we test all the models on raw images and those combined with perturbations.
We observe our method reaches to the competing performance comparable to the other state of the arts. In particular, our detector using stage 5 artifacts is the most stable against perturbations, which demonstrates the effectiveness and necessity of involving high-level information. For example, our detectors (i.e., SR+C and SR+D) achieve more than 95\% accuracy on church\_outdoor with perturbations while others are less than 70\%.
We conclude the proposed method is able to cope with diverse challenges using multi-level features and our re-synthesis module. It sheds the light on the new way for fake image detection, more advantageous to only analyzing the local patterns~\cite{marra2019fingerprints,ning2019iccv_gan_detection} or frequency distributions~\cite{frank2020dct2d_detect,zhang2019detecting}. {Additionally, we show the class activation maps for LSUN in Figure~\ref{fig:cam_artifacts_bedroom}. For facial images, we observe our methods locate fake regions more on background, which is actually harder to generate than faces. Different from facial images, in LSUN, our CAMs are more flexible in highlighted shapes and locations for highly diverse scene images.}

\section{Conclusion}
\label{sec:conclusion}
Due to the unsustainable reliance of prior work on frequency artifacts, we show how these methods deteriorate when such artifacts are suppressed. Based on this insight, we present a novel feature representation for detecting {Deepfakes}. Instead of the limited focus on low-level local artifact characteristics, we reason that different levels of information can help detect fake images with beneficial synergy. The hierarchical artifacts from a re-synthesizer are evidenced to boost the performance, generalization, and robustness of a downstream detector. These have been validated with {high-resolution Deepfakes} created from state-of-the-art GANs. 
{Further, deepfakes detection is still an open problem because many issues are unresolved.  Attacks can reverse the procedure of classification to fool a detector, therefore, we believe our solution provides a promising solution that our features are from a parameterized model which can be watermarked, and thus increases the challenges of attacks on our classifier.}
In the end, we conclude deepfakes detection requires more than analysis {in addition to} low-level details, but also on higher-level visual cues which has the potential to lead to more sustainable detection schemes.

\section*{Acknowledgements}
Ning Yu is partially supported by the Twitch Research Fellowship.

{
\bibliographystyle{named}
\bibliography{ijcai21}
}

\end{document}